\documentclass[journal,comsoc]{IEEEtran}
\usepackage{adjustbox}
\usepackage[english]{babel}

\usepackage{graphicx}
\usepackage[utf8]{inputenc}
\usepackage{booktabs} 
\usepackage{blindtext, graphicx,subfig, mathtools,amsmath,float,url,textcomp,multirow, multirow,pgfplots,textcomp,adjustbox,adjustbox,array,tikz}
\usetikzlibrary{patterns}
\usepgfplotslibrary{groupplots}
\pgfplotsset{compat=1.7}
\usepackage{xcolor}
\usepackage{array}
\usepackage{algorithm}
 \newcommand{\alginput}{\textbf{Input: }}
 \usepackage{color, colortbl}
 \definecolor{LightCyan}{rgb}{0.88,0.86,0.88}
 \usepackage{xcolor}
\usepackage{algpseudocode}%
\usepackage{multirow}
\usepackage{balance}
\usepackage[T1]{fontenc}

\ifCLASSINFOpdf
\else
\fi
\usepackage{amsmath}
\usepackage[cmintegrals]{newtxmath}
\begin{document}
%
\title{ASD-DiagNet: A hybrid learning approach for detection of Autism Spectrum Disorder using fMRI data}
%
%
%
\author{Taban~Eslami,
        Vahid~Mirjalili, 
        Alvis~Fong,
        Angela~Laird,
        and~Fahad~Saeed$^*$ 
}

\author{Taban~Eslami,
        Vahid~Mirjalili,
        Alvis~Fong,
         Angela~Laird,
        and~Fahad~Saeed$^*$,
\thanks{T. Eslami and A. Fong are with the Department
of Computer Science, Western Mihcigan university, Kalamazoo,
MI, 49008. E-mail: {taban.eslami,alvis.fong}@wmich.edu}
\thanks{V. Mirjalili is with Department of Computer Science and Engineering, Michigan State University, Lansing,
MI, 48824. E-mail: {mirjalil}@msu.edu}
\thanks{A. Laird is with Department of Physics, Florida International University, Miami,
FL, 33199. E-mail: {alaird}@fiu.edu}
\thanks{F. Saeed is with school of computing and information science, Florida International University, Miami,
FL, 33199. $^*$ Corresponding E-mail: {fsaeed}@fiu.edu}
}

%
%

\markboth{}%
{Shell \MakeLowercase{\textit{et al.}}: }
%



\maketitle

\begin{abstract}
Mental disorders such as Autism Spectrum Disorders (ASD) are heterogeneous disorders that are notoriously difficult to diagnose, especially in children. The current psychiatric diagnostic process is based purely on the behavioural observation of symptomology (DSM-5/ICD-10) and may be prone to over-prescribing of drugs due to misdiagnosis. In order to move the field towards more quantitative fashion, we need advanced and scalable machine learning infrastructure that will allow us to identify reliable biomarkers of mental health disorders. In this paper, we propose a framework called ASD-DiagNet for classifying subjects with ASD from healthy subjects by using \emph{only} fMRI data. We designed and implemented a joint learning procedure using an autoencoder and a single layer perceptron which results in improved quality of extracted features and optimized parameters for the model. Further, we designed and implemented a data augmentation strategy, based on linear interpolation on available feature vectors, that allows us to produce synthetic datasets needed for training of machine learning models. The proposed approach is evaluated on a public dataset provided by Autism Brain Imaging Data Exchange including 1035 subjects coming from 17 different brain imaging centers. Our machine learning model outperforms other state of the art methods from 13 imaging centers with increase in classification accuracy up to 20\% with maximum accuracy of 80\%. The machine learning technique presented in this paper, in addition to yielding better quality, gives enormous advantages in terms of execution time (40 minutes vs. 6 hours on other methods). The implemented code is available as GPL license on GitHub portal of our lab (https://github.com/pcdslab/ASD-DiagNet).
\end{abstract}


%
\IEEEpeerreviewmaketitle

\section{Introduction}
%
%
%
%

\IEEEPARstart{M}{ental} disorders such as Autism Spectrum Disorders (ASD) are heterogeneous disorders that are notoriously difficult to diagnose, especially in children. The current psychiatric diagnostic process is based purely on behavioural observation of symptomology (DSM-5/ICD-10) and may be prone to misdiagnosis \cite{nickel2017early}. There is no quantitative test that can be prescribed to patients that may lead to definite diagnosis of a person. Such quantitative and definitive tests are a regular practice for other diseases such as diabetes, HIV, and hepatitis-C. It is widely known that defining and diagnosing mental health disorders is a difficult process due to overlapping nature of symptoms, and lack of a biological test that can serve as a definite and quantified gold standard \cite{national2018attention}. Autism Spectrum Disorders (ASD) is a lifelong neuro-developmental brain disorder which causes social impairments like repetitive behaviour and communication problems in children. More than $1\%$ of children suffer from this disorder and detecting it at early ages can be beneficial. Studies show that some demographic attributes like gender and race vary among ASD and healthy individuals such that males are four times more prone to ASD than females~\cite{baio2018prevalence}.

Quantitative analysis of brain imaging data can provide valuable biomarkers that result in more accurate diagnosis of brain diseases. Machine learning techniques using brain imaging data (e.g. Magnetic Resonance Imaging (MRI) and functional Magnetic Resonance Imaging (fMRI)) have been extensively used by researchers for diagnosing brain disorders like Alzheimer's, ADHD, MCI and, Autism.~\cite{hosseini2016alzheimer,eslami2018sing,khazaee2017classification,yang2014deep,peng2013extreme,colby2012insights,deshpande2015fully}.

In this paper, we focus on classifying subjects suffering from Autism Spectrum Disorders (ASD) from healthy control subjects using fMRI data. We propose a method called \emph{ASD-DiagNet} which consists of an autoencoder and a single layer perceptron. These networks are used for extracting lower dimensional features in a hybrid manner and the trained perceptron is used for the final round of classification. In order to enlarge the size of the training set, we designed a data augmentation technique which generates new data in feature space by using available data in the training set. Based on the experimental results, ASD-DiagNet achieved $70.1\%$ classification accuracy which outperforms the current state of the art technique~\cite{heinsfeld2018identification}. 
Further, we show that ASD-DiagNet scales extremely well with increasing size of the data and takes only 41 minutes to run as compared to $6$ hours needed by other methods \cite{heinsfeld2018identification}. Average accuracy on individual sites is $63\%$, which is $7\%$ better than the result reported by~\cite{heinsfeld2018identification}. Our machine learning technique will allow greater quantification of ASD diagnosis and is a step forward to making the early diagnosis and treatment a priority.

The structure of this paper is as follows: In the next section, we explain the state of the art in the field. In Section~\ref{sec:material}, we explain ASD-DiagNet method in detail. In Section~\ref{sec:experiments}, we describe the experiment setting and discuss the results of ASD-DiagNet. Finally, in Section~\ref{sec:conclusion}, we conclude the paper and discuss future direction.
\section{Background Information and Literature Review}\label{sec:Background}

 Detecting ASD using fMRI data has recently gained a lot of attention, thanks to Autism Brain Imaging Data Exchange (ABIDE) initiative for providing functional and structural brain imaging datasets collected from several brain imaging centers around the world~\cite{craddock2013neuro}. Many studies and methods have been developed based on ABIDE data~\cite{heinsfeld2018identification,iidaka2015resting,chen2016multivariate,abraham2017deriving}.
 Some studies included a subset of this dataset based on specific demographic information to analyze their proposed method. For example, Iidaka~\cite{iidaka2015resting} used probabilistic neural network for classifying resting state fMRI (rs-fMRI) data from $312$ ASD and $328$ healthy control subjects (Subjects under $20$ years old were selected) which achieved around $90\%$ accuracy. In another work, Plit et al.~\cite{plitt2015functional} used two sets of rs-fMRI data, one containing $118$ male individuals ($59$ ASD; $59$ TD) and the other containing $178$ age and IQ matched individuals ($89$ ASD; $89$ TD) from ABIDE dataset and achieved $76.67\%$ accuracy. 
 
 Besides using fMRI data, some studies also included structural and demographic information of subjects for diagnosing ASD. Parisot et al.~\cite{parisot2018disease} proposed a framework based on Graph Convolutional Networks that achieved $70.4\%$ accuracy. In their work, they represented the population as a graph in which nodes are defined based on imaging features and phenotypic information describe the edge weights. Sen et al.~\cite{sen2018general} proposed a new algorithm which combines structural and functional features from MRI and fMRI data and got $64.3\%$ accuracy by using $1111$ total healthy and ASD subjects. Nielsen et al.~\cite{nielsen2013multisite} obtained $60\%$ accuracy on a group of $964$ healthy and ASD subjects using the functional connectivity between 7266 regions and demographic information like age, gender, and handedness attributes. 
 
 Machine learning techniques such as Support Vector Machines (SVM) and Random Forests are explored in multiple studies~\cite{subbaraju2017identifying,fredo2018diagnostic,abraham2017deriving,bi2018classification}. For instance, Chen et al.~\cite{chen2016multivariate} investigated the effect of different frequency bands for constructing brain functional network, and obtained $79.17\%$ accuracy using SVM technique applied to $112$ ASD and $128$ healthy control subjects.
 
 Recently, using neural networks and deep learning methods such as autoencoders, Deep Neural Network (DNN), Long Short Term Memory (LSTM) and Convolutional Neural Network (CNN) have also become very popular for diagnosing ASD~\cite{guo2017diagnosing,bi2018diagnosis,brown2018connectome,dvornek2017identifying,li2018novel,khosla20183d}. Brown et al.~\cite{brown2018connectome} obtained $68.7\%$ classification accuracy on $1013$ subjects composed of $539$ healthy control and $474$ with ASD, by proposing an element-wise layer for deep neural networks which incorporated the data-driven structural priors. 
 
 Most recently, Heinsfeld et al.~\cite{heinsfeld2018identification} used a deep learning based approach and achieved $70\%$ accuracy for classifying $1035$ subjects ($505$ ASD and $530$ controls). They claimed this approach improved the state of the art technique. In their technique, distinct pairwise Pearson's correlation coefficients were considered as features. Two stacked denoising autoencoders were first pre-trained in order to extract lower dimensional data. After training autoencoders, their weights were applied to a multi-layer perceptron classifier (fine-tuning process) which was used for the final classification. 
 However, they also performed classification for each of the $17$ sites included in ABIDE dataset separately, and the average accuracy is reported as $52\%$. The low performance on individual sites was justified to be due to the lack of enough training samples for intra-site training. 
 
 Generally, most related studies for ASD diagnosis using machine learning techniques have only considered a subset of ABIDE dataset, or they have incorporated other information besides fMRI data in their model.
 There are few studies such as \cite{heinsfeld2018identification}, which only used fMRI data without any assumption on demographic information and analyzed \textit{all} the $1035$ subjects in ABIDE dataset. To the best of our knowledge~\cite{heinsfeld2018identification} is currently state of the art technique for ASD diagnosis on whole ABIDE dataset, which we use as the baseline for evaluating our proposed method.
 
 \section{Materials and methods}\label{sec:material}
\subsection{Functional Magnetic Resonance Imaging and ABIDE dataset}
Functional Magnetic Resonance Imaging (fMRI) is a brain imaging technique that is used for studying brain activities \cite{lindquist2008statistical,eslami2018fast}. In fMRI data, the brain volume is represented by a group of small cubic elements called voxels. A time series is extracted from each voxel by keeping track of its activity over time. Scanning the brain using fMRI technology while the subject is resting is called resting state fMRI (rs-fMRI), which is widely used for analyzing brain disorders.
In this study, we used preprocessed ABIDE-I dataset that is provided by the ABIDE initiative. This dataset consists of $1112$ rs-fMRI data including ASD and healthy subjects collected from $17$ different sites. We used fMRI data of the same group of subjects which was used in \cite{heinsfeld2018identification}. This set consists of $505$ subjects with ASD and $530$ healthy control from all the $17$ sites. Table \ref{fig:classmembership} shows the class membership information for each site.
\begin{table*}
\centering
\noindent\setlength\tabcolsep{3pt}
\caption{Class membership information of ABIDE-I dataset for each individual site}
\label{fig:classmembership}
\scalebox{1}{
\begin{tabular}{|c|c c c c c c c c c c c c c c c c c |}
 \hline
 Site& Caltech & CMU & KKI & Leuven & MaxMun & NYU & OHSU & OLIN & PITT & SBL & SDSU & Stanford & Trinity & UCLA & UM & USM & Yale \\
 \hline
 ASD& 19 & 14 & 20 & 29 & 24 & 75 & 12 & 19 & 29 & 15 & 14 & 19 & 22 & 54 & 66 & 46 &28 \\
Healthy control& 18 & 13 & 28 & 34 & 28 & 100 & 14 & 15 & 27 & 15 & 22 & 20 & 25 & 44 & 74 & 25 & 28 
 \\  \hline
\end{tabular}
}
\end{table*}
ABIDE-I provided the average time series extracted from seven sets of regions of interest (ROIs) based on seven different atlases which are preprocessed using four different pipelines. The data used in our experiments is preprocessed using C-PAC pipeline~\cite{craddock2013neuro} and is parcellated into $200$ functionally homogeneous regions generated using spatially constrained spectral clustering algorithm~\cite{craddock2012whole} (CC-200). The preprocessing steps include slice time correction, motion correction, nuisance signal removal, low frequency drifts and voxel intensity normalization. It is worth mentioning that each site used different parameters and protocols for scanning the data. Parameters like 
repetition time (TR), echo time (TE), number of voxels, number of volumes, openness or closeness of the eyes while scanning are different among sites.

\subsection{ASD-DiagNet: Feature extraction and classification}\label{sec:asd-diagnet_model}

\begin{figure*}[h]
\centering
\includegraphics[width=13 cm,height=5 cm]{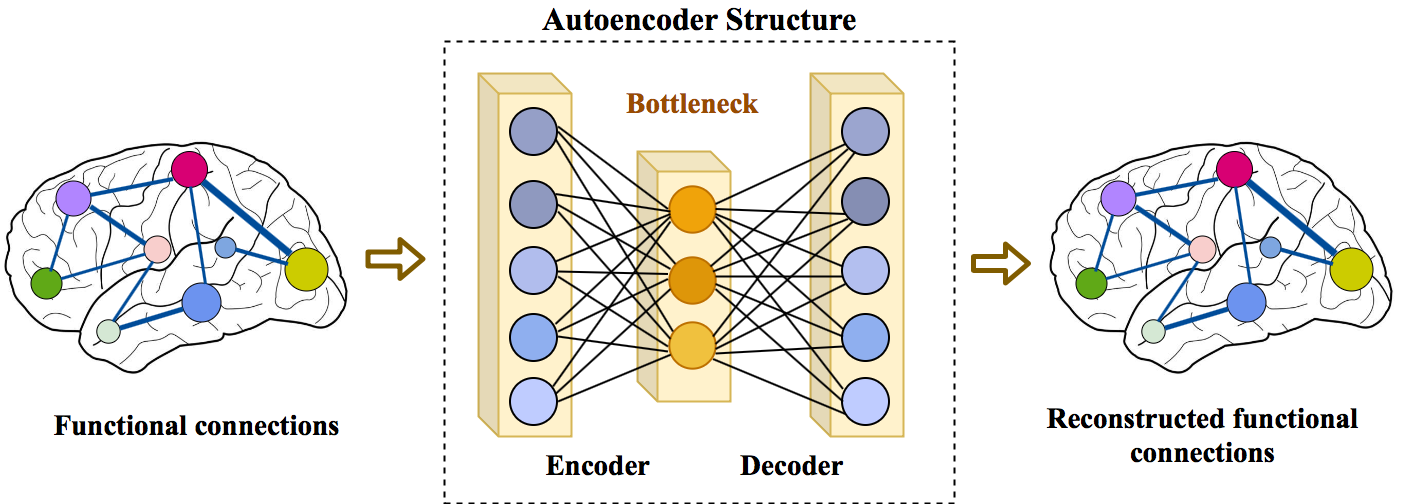}
\caption{Structure of an autoencoder consisting of an encoder that receives the input data and encodes it into a lower dimensional representation at the bottleneck layer, and a decoder that reconstructs the original input from the bottleneck layer.}
\label{fig:autoencoder}
\end{figure*}

Functional connectivity between brain regions is an important concept in fMRI analysis and is shown to contain discriminative patterns for fMRI classification. Among correlation measures, Pearson's correlation is mostly used for approximating the functional connectivity in fMRI data ~\cite{liang2012effects,zhang2017hybrid,baggio2014functional}. It shows the linear relationship between the time series of two different regions. Given two times series, $u$ and $v$, each of length $T$, the Pearson's correlation can be computed using the following equation:
\begin{equation}
    \rho_{uv}= \dfrac{\sum_{t=1}^{T} (u_t - \bar{u})(v_t - \bar{v})}{\sqrt{\sum_{t=1}^{T} (u_t - \bar{u})^2}\sqrt{\sum_{t=1}^{T} (v_t - \bar{v})^2}}
\end{equation}
where $\bar{u}$ and $\bar{v}$ are the mean of times series $u$ and $v$, respectively.  
Computing all pairwise correlations results in a correlation matrix $\mathcal{C}_{m\times m}$ where $m$ is the number of time series (or regions). Due to the symmetric property of Pearson's correlation, we only considered the strictly upper triangle part of the correlation matrix. Since we used CC-200 atlas in which the brain is parcellated into $m=200$ regions, there are $m\times(m-1)/2=19900$ distinct pairwise Pearson's correlations. 
In this regard, we selected half of the correlations comprising $1/4$ largest and $1/4$ smallest values and eliminated the rest. To do so, we first compute the average of correlations among all subjects in training set and then pick the indices of the largest positive and negative values from averaged correlation array. We then pick the correlations at those indices from each sample as our feature vector. Keeping half of the correlations and eliminating the rest reduces the size of input features by a factor of 2. There is no limitation of the number of high- and anti-correlations that should be kept. Removing more features results in higher computational efficiency as well as reducing the chance of overfitting, however removing too many features can also cause loosing important patterns. 

In order to further reduce the size of features, we used an autoencoder to extract a lower dimensional feature representation. An autoencoder is a type of feed-forward neural network model, which first encodes its input $x$ to a lower dimensional representation,
\begin{equation}
h_{enc}=\phi_{enc}(x) = \tau \left(W_{enc}x + b_{enc}\right)    
\end{equation}
where $\tau$ is the hyperbolic tangent activation function ($Tanh$), and $W_{enc}$ and $b_{enc}$ represent the weight matrix and the bias for the encoder. Then, the decoder reconstructs the original input data
\begin{equation}
x^\prime=\phi_{dec}(h_{enc}) = W_{dec}h_{enc} + b_{dec}
\end{equation}
where $W_{dec}$ and $b_{dec}$ are the weight matrix and bias for the decoder. In this work, we have designed an autoencoder with tied weights, which means $W_{dec} = W_{enc}^\top$. 
An autoencoder can be trained to minimize its reconstruction error, computed as the Mean Squared Error (MSE) between $x$ and its reconstruction, $x^\prime$. The choice of using autoencoder instead of other feature extraction techniques like PCA is its ability to reduce the dimensionality of features in a non-linear way. Structure of an autoencoder is shown in Fig.~\ref{fig:autoencoder}.

The lower dimensional data generated during the encoding process contains useful patterns from the original input data with smaller size, and can be used as new features for classification.
For the classification task, we used a single layer perceptron (SLP) which uses the bottleneck layer of the autoencoder, $h_{enc}$, as input, and computes the probability of a sample belonging to the ASD patient class using a sigmoid activation function, $\sigma$,
\begin{equation}
\begin{array}{rl}
    f(x) & = \sigma \left(W_{slp}h_{enc} + b_{slp}\right)  \\
     & = \sigma \left(W_{slp} \tau(W_{enc}x + b_{enc}) + b_{slp}\right)
\end{array}
\end{equation}
where $W_{slp}$ and $b_{slp}$ are the weight matrix and the bias for the SLP network. The SLP network can be trained by minimizing the Binary Cross Entropy loss, $\mathcal{H}$, using the ground-truth class label, $y$, and the estimated ASD probability for each sample, $f(x)$:
\begin{equation}
    \mathcal{H}(y, f(x)) = -\left(y\times f(x) + (1-y)\times(1-f(x))\right) 
\end{equation}
Finally, the predicted class label is determined by thresholding the estimated probability

\begin{equation}
  \hat{y}=\begin{cases}
    1, & \text{if }f(x)\ge 0.5,\\
    0, & \text{otherwise}.
  \end{cases}
\end{equation}

Typically, an autoencoder is fully trained such that its reconstruction error is minimized, then, the features from bottleneck layer, $h_{enc}$, are used as input for training the SLP classifier, separately.
In contrast, here, we train the autoencoder and the SLP classifier simultaneously. This can potentially result in obtaining low dimensional features that have two properties 
\begin{enumerate}
    \item useful for reconstructing the original data,
    \item contain discriminative information for the classification task.
\end{enumerate}
This is accomplished by adding the two loss functions, i.e. MSE loss for reconstruction, and Binary Cross Entropy for the classification task, and training both networks jointly.
After the joint training process is completed, we further fine-tune the SLP network for a few additional epochs, while parameters of the autoencoder are frozen.
\subsection{Data augmentation using linear interpolation}

\begin{figure*}[ht]
\centering
\includegraphics[width=\textwidth ,height=30 cm,keepaspectratio]{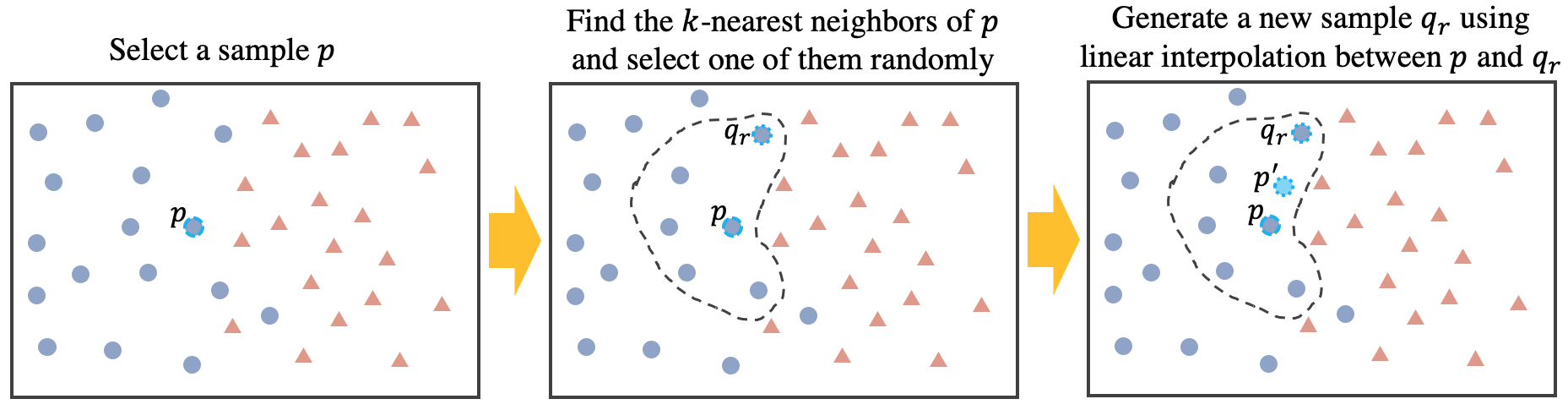}
\caption{Generating new artificial data: Step 1) Selecting a sample ($p$). Step 2) Find $k$-nearest neighbors of $p$ from the same class, and pick one random neighbor ($q_r$). 3) Generate new sample $p'$ using $p$ and $q_r$ by linear interpolation. 
}
\label{fig:augmentation}
\end{figure*}

\begin{figure*}[h]
\centering
\includegraphics[width=\textwidth,height=15 cm,keepaspectratio]{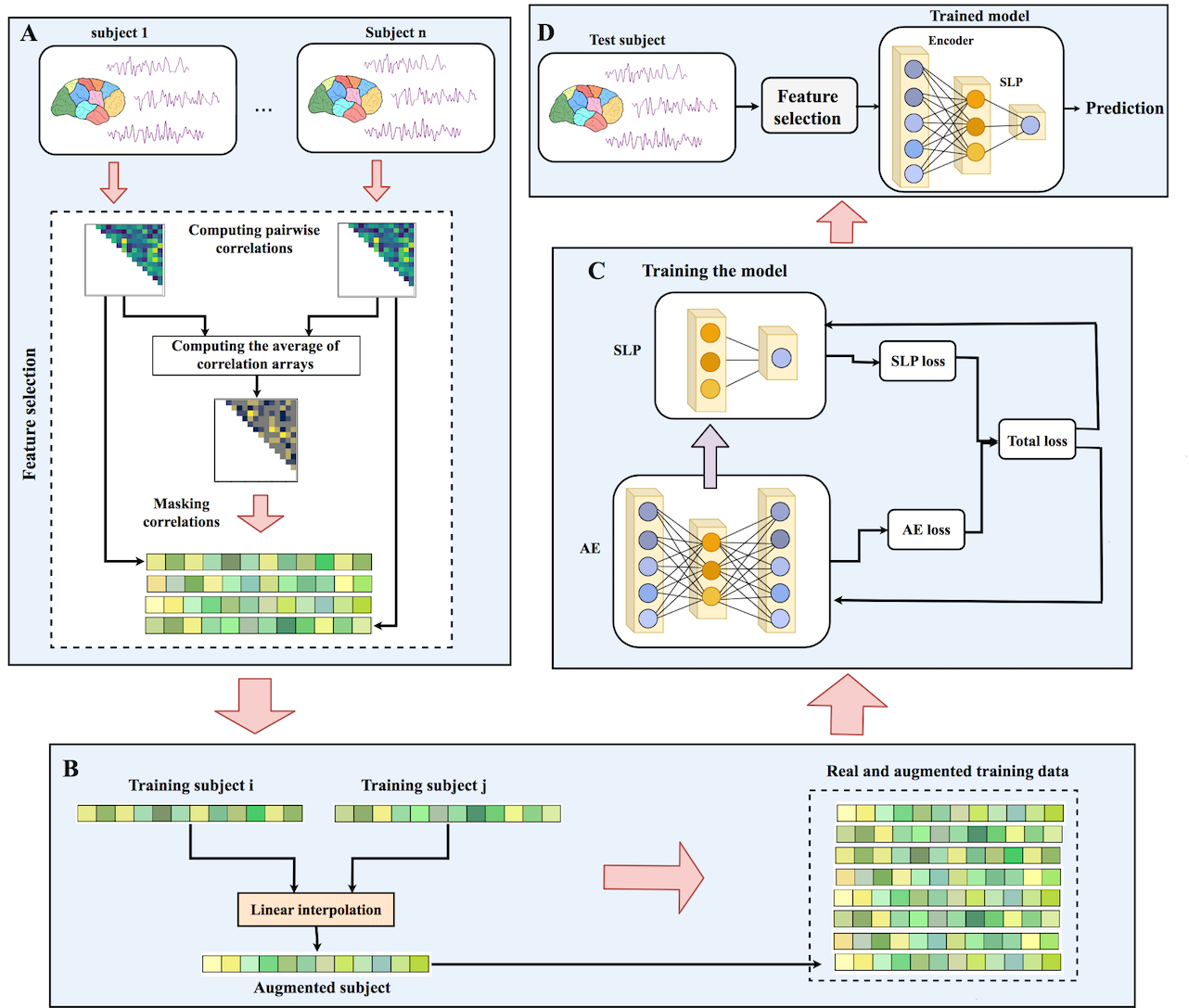}
\caption{ Workflow of ASD-DiagNet: A) Pairwise Pearson's correlations for each subject in the training set is computed.
The average of all correlation arrays is computed and the position of $1/4$ largest and $1/4$ smallest values in the average array is considered as a mask. Masked correlation array of each sample is considered as its feature vectors. B) A set of artificial samples is generated using the feature vectors of training samples. C) Autoencoder and SLP are jointly trained by adding up their training loss in each iteration. D) For a test subject, the features are extracted using the mask generated in part A, followed by passing the features through the encoder part of the autoencoder, and finally predicting its label using the trained SLP. 
}
\label{fig:framework}
\end{figure*}
Machine learning and especially deep learning techniques can be advantageous if they are provided with enough training data. Insufficient data causes overfitting and non-generalizability of the model~\cite{raschka2017python}. Large training sets are not always available and collecting new data might be costly like in medical imaging field. In these situations, data augmentation techniques can be used for generating synthetic data using the available training set \cite{wong2016understanding,perez2017effectiveness,eitel2015multimodal,karpathy2014large,xu2016improved}. The data augmentation technique that we propose in this study is inspired by Synthetic Minority Over-sampling Technique (SMOTE)~\cite{chawla2002smote}. SMOTE is an effective model which is used for oversampling the data in minority class of imbalanced datasets.
 SMOTE generates synthetic data in feature space by using the nearest neighbors of a sample. After k-nearest neighbors of sample $p$ are found ($\{q_1, q_2, ..., q_k\}$), a random neighbor is selected ($q_r$) and the synthetic feature vector is computed using the following equation:
 \begin{equation}
    p^\prime=\alpha \times p+(1-\alpha)\times q_r
 \end{equation}
 In this equation, $\alpha$ is a random number selected uniformly in the range $[0, 1]$. In our implementation, we chose $\alpha$ randomly within range $[0.5, 1]$, so that the synthesized sample is closer to $p$. 
 Finding the nearest neighbors of a sample is based on a distance or similarity metric. In our work, the samples have feature vectors of size $9950$ (half of the correlations). One idea for computing nearest neighbors is to use Euclidean distance, however, computing the pairwise Euclidean distances with $9950$ features is not efficient. 
 In order to compute the similarity between samples and finding the nearest neighbors, we used a measure called Extended Frobenius Norm (EROS). This measure computes the similarity between two multivariate time series (MTS)~\cite{yang2004pca}. 
 fMRI data consists of several regions each having a time series so we can consider it as a multivariate time series. Our previous study on ADHD disorder has shown that EROS is an effective similarity measure for fMRI data and using it along with k-Nearest-Neighbor achieves high classification accuracy \cite{eslami2018sing}. This motivated us to utilize it as part of the data augmentation process. EROS computes the similarities between two MTS items $A$ and $B$ based on eigenvalues and eigenvectors of their covariance matrices using the following equation:
 \begin{equation}
 \begin{array}{rl}
 EROS(A,B,w) &= \sum_{i=1}^{n}{w_i \left|\langle a_i,b_i \rangle \right|} \\& = \sum_{i=1}^{n}{w_i\left|cos \theta_i \right|}
 \end{array}
 \end{equation}
where, $\theta_i$ is the cosine of the angle between $i_{th}$ corresponding eigenvectors of covariance matrices of multivariate time series $A$ and $B$. Furthermore, $w$ is the weight vector which is computed based on eigenvalues of all MTS items using Algorithm~\ref{alg:EROS}.
This algorithm computes the weight vector $w$ by normalizing eigenvalues of each MTS item followed by applying an aggregate function $f$ (here, we used mean) to all eigenvalues over the entire training dataset and finally normalizing them so that $\sum_{i=1}^{n} w_i = 1$.
  
 \begin{algorithm}
  \caption{Computing weight vector for EROS~\cite{yang2004pca}}
\label{alg:EROS}
 \begin{flushleft}
  \alginput  
  An $n\times N$ matrix $S$, where $n$ is the number of variables for the dataset and $N$ is the number of MTS items in the dataset. Each column vector $s_i$ in $S$ represents all the eigenvalues for $i_{th}$ MTS item in the dataset. $s_{ij}$ is a value at column $i$ and row $j$ in $S$. $s_{*i}$ is $i_{th}$ row in $S$. $s_{i*}$ is $i_{th}$ column
  \end{flushleft}
  \begin{algorithmic}[1]
   \For {$i=1$ to $N$} 
  \State {$s_i \leftarrow s_i/\ \sum_{j=1}^{n} s_{ij}$}
  \EndFor
  \For {$i=1$ to $n$} 
  \State {$w_i \leftarrow f(s_{*i})$}
  \EndFor
  \For {$i=1$ to $n$} 
  \State {$w_i \leftarrow w_i/\sum_{j=1}^{n} w_j$}
  \EndFor
  \end{algorithmic}
\end{algorithm}

In order to further reduce the time needed for computing the pairwise similarities, we considered using the first two eigenvectors of each sample. Our experiments showed that this simplification does not affect the results while reducing the running time significantly compared to using all eigenvectors and eigenvalues. 

Now, using EROS as the similarity measure, our data augmentation process is shown in Algorithm~\ref{alg:augEROS}. After finding $k=5$ nearest neighbors of each sample $i$ in the training set, one of them is randomly selected, a new sample is generated using linear interpolation between the selected neighbor and sample $i$. Using this approach, one synthetic sample is created for each training point which results in doubling the size of the training set. Fig.~\ref{fig:augmentation} shows the data augmentation process and Fig.~\ref{fig:framework} shows the overall process of ASD-DiagNet method.
\begin{algorithm}
  \caption{Data augmentation using EROS similarity measure}
\label{alg:augEROS}
 \begin{flushleft}
  \alginput  
  Training dataset of size N
  \end{flushleft}
  \begin{algorithmic}[1]
   \For {$i=1$ to $N$} 
  \State {Find $5$ nearest neighbors to $i$ using EROS}
  \State{$j \leftarrow$ A random sample among nearest neighbors}
  \State {$r \leftarrow$ Random number in the range $[0.5, 1]$ }
  \State {$x_{i+N}^{*} \leftarrow \alpha \times x_i + (1-\alpha) \times x_j$ }
  \EndFor
 
  \end{algorithmic}
\end{algorithm}

\section{experiments and results}\label{sec:experiments}
 
 For all the experiments reported in this section, we used a Linux server running Ubuntu Operating System. The server contains two Intel Xeon E5-2620 Processors at $2.40$ GHz with a total $48$ GBs of RAM. The system contains an NVIDIA Tesla K-40c GPU with $2880$ CUDA cores and $12$ GBs of RAM. CUDA version $8$ and PyTorch library were used for conducting the experiments.
 
We evaluated ASD-DiagNet model in two phases. In the first phase, the model was evaluated using the whole $1035$ subjects from all sites and in the second phase, the model was evaluated for each site separately. As stated earlier, data centers may have used different experimental parameters for scanning fMRI images, so considering all of them in the same pool determines how our model generalizes to data with heterogeneous scanning parameters. On the other hand, by considering each data center separately, fewer subjects are available for training the model and the results indicate how it performs on small datasets. In each of these experiments, the effect of data augmentation was evaluated.
The following subsections explain each experiment in more details. 
\subsection{Phase 1: Experiments using the whole dataset}

In this phase, we performed 10-fold cross-validation on the whole $1035$ subjects.
Table \ref{table:acc_overal} compares accuracy, sensitivity, and specificity of our approach with the method proposed by Heinsfeld et al.~ \cite{heinsfeld2018identification}, random forest, and SVM with RBF kernel classifier. SVM and random forest were trained using $19900$ pairwise Pearson's correlations for each subject.  As the results show, ASD-DiagNet achieves $70.1\%$ which outperforms other methods. The proposed data augmentation helps to improve the results by around $1\%$.\footnote{We like to mention that Heinsfeild~\cite{heinsfeld2018identification} reported $70\%$ accuracy in their paper, however, the accuracy we reported here is the result of running their method on our system using their default parameters and the code they provided online. The different results observed here could be due to some missing details in the implementation.}

\begin{table}[h]
\centering
\noindent\setlength\tabcolsep{2.5pt}
\caption{Classification performance using 10-fold cross-validation on the whole dataset; Note that our proposed approach, ASD-DiagNet (with data augmentation) achieves highest accuracy among existing methods.}
\label{table:acc_overal}
\begin{tabular}{|c|c c c|}
 \hline
 Method& Accuracy & Sensitivity & Specificity \\
 \hline
 
  ASD-DiagNet & 70.1 & 67.8 & 72.8 \\
  ASD-DiagNet (no aug.) & 69.2 & 66.4 & 73.1 \\
SVM & 60.3 & 35 & 84.4\\
Random Forest & 63 & 54.9 & 71.3\\
Heinsfeild et al.~\cite{heinsfeld2018identification}& 65.4 & 69.3 & 61 \\  \hline
\end{tabular}
\end{table}
\subsection{Phase 2: Intra-site evaluation}

In this phase, we performed 5-Fold cross-validation on each site, separately. The accuracy of each method is provided in Table~\ref{table:IntraSite}. 
Based on these results, our method achieves the highest accuracy in most cases and outperforms other methods on average.
In addition, note that the proposed data augmentation helps improving the result around $2\%$ overall. Especially, for OHSU, the data augmentation improves the accuracy significantly ($15\%$ increase).

\begin{table}[h]
\centering
\noindent\setlength\tabcolsep{2.5pt}
\caption{Classification accuracy using 5-fold cross-validation on individual data centers using our proposed method, ASD-DiagNet (with and without data augmentation), compared with other existing methods.}
\label{table:IntraSite}
\centering
\begin{tabular}{|c| c | c | c | c | c|}
 \hline
 \multirow{2}{*}{Site}& \multirow{2}{*}{ASD-DiagNet} & ASD-DiagNet &
  \multirow{2}{*}{Ref.~\cite{heinsfeld2018identification}}& \multirow{2}{*}{SVM} & Random- \\ 
  & & (no aug.) & & & Forest\\\hline \hline
   Caltech & 51.4 & 49.2 & 52.3 & 48.5 & \cellcolor{blue!25}\textbf{55.4}
 \\ 
  CMU & 63.6 & 62.5 & 45.3 & 60 & \cellcolor{blue!25}\textbf{64.6}
 \\  
  KKI & \cellcolor{blue!25}\textbf{70.6} & 66.6 & 58.2 & 58.2 & 67.6
 \\  
  Leuven & \cellcolor{blue!25}\textbf{59} & 57.2 & 51.8 & 53.9 & 57.5
 \\  
  MaxMun  & 48.3 & 48 & \cellcolor{blue!25}\textbf{54.3} & 53.8 & 45.8
 \\ 
 NYU & \cellcolor{blue!25}\textbf{68.5} & 66.1 & 64.5 & 57.1 & 62.3
 \\  
  \textbf{OHSU} & \cellcolor{blue!25}\textbf{\textcolor{red}{80}} & 65.33 & 74 & 54 & 54.4
 \\
 Olin & \cellcolor{blue!25}\textbf{64.7} & 61.33 & 44 & 55.7 & 53.4
 \\ 
  Pitt & \cellcolor{blue!25}\textbf{68} & 66.8 & 59.8 & 51.8 & 60.87
 \\ 
 SBL & \cellcolor{blue!25}\textbf{53} & 52.3 & 46.6 & 50 & 47.6
 \\  
 SDSU & \cellcolor{blue!25}\textbf{63.9} & 63 & 63.6 & 61.1 & 61.9
 \\  
  Stanford & \cellcolor{blue!25}\textbf{62.5} & 61.5 & 48.5 & 51.4 & 60.1
 \\ 
 Trinity & 52.9 & 53.3 & \cellcolor{blue!25}\textbf{61} & 53.3 & 52.6
 \\ 
 UCLA & \cellcolor{blue!25}\textbf{72} & 71.3 & 57.7 & 55.1 & 69.3
 \\  
 USM & \cellcolor{blue!25}\textbf{69} & 64 & 62 & 64.7 & 64.7
 \\  
 UM & 64.2 & \cellcolor{blue!25}\textbf{64.7} & 57.6 & 52.8 & 63.5
 \\  
 Yale & \cellcolor{blue!25}\textbf{63.2} & 61.3 & 53 & 57.6 & 58.2
 \\  \hline
 Average & \cellcolor{blue!25}\textbf{63.2} & 60.8 & 56.1 & 55.1 & 59.8
 \\  \hline
\end{tabular}
\end{table}
\subsection{Running time} 
The running time needed for performing 10-fold cross-validation by different approaches is shown in Table \ref{table:running_time}. The training and evaluation for all methods are performed on the same Linux system (described in Section~\ref{sec:experiments}).
\begin{table}[H]
\noindent\setlength\tabcolsep{4.0pt}
\caption{Running time for 10-fold cross-validation (training and evaluation) on the whole dataset.}
\label{table:running_time}
\centering
\begin{tabular}{|c|c|}
 \hline
 Method& Running time\\
 \hline
  ASD-DiagNet & $41.14$ min\\
  ASD-DiagNet (no aug.) & $20.5$ min\\
SVM & $3$ min\\
Random forest & $1$ min\\
Heinsfeild et al~\cite{heinsfeld2018identification}& $6$ hr\\
 \hline
\end{tabular}
\end{table}

Based on the results in Table~\ref{table:running_time}, ASD-DiagNet performs significantly faster than~\cite{heinsfeld2018identification}. The data augmentation doubles the size of the training set by generating one artificial sample per subject in the training set. As a result, the data augmentation increases the computation time by a factor of $2$. 

\subsection{Experiment on other parcellations}

We tested ASD-DiagNet on two other ROI atlases besides CC-200. The first parcellation is based on Automated Anatomical Labeling (AAL) atlas in which the brain is parcellated into 116 regions using AAL toolbox. The other atlas is called Dosenbach160 which parcellates the brain into 160 regions. The data for these parcellations is also provided in ABIDE dataset.
Dosenbach160 and AAL contain 12720 and 6670 pairwise correlations, respectively. Similar to CC-200 atlas, half of the correlations (keeping the 1/4 largest and 1/4 smallest values, and removing the rest intermediate values) are selected as input features to the model.  
 The resulting average accuracy, sensitivity, and specificity of performing 10-fold cross-validation on the whole dataset using different approaches for AAL and Dosenbakh160 are shown in Table~\ref{table:acc_otherregions}.

\begin{table}[H]
\centering
\noindent\setlength\tabcolsep{2.5pt}
\caption{Classification accuracy using other parcellations of brain fMRI data: AAL and Dosenbach160; Note that our proposed method, ASD-DiagNet, outperforms existing techniques using both atlases.}
\label{table:acc_otherregions}
\begin{tabular}{|c|c c|}
 \hline
 Method & AAL & Dosenbach160  \\
 \hline
  ASD-DiagNet & \textbf{67.8} & \textbf{65}
\\
  ASD-DiagNet (no augmentation) & 65.6 & 64.3 
\\
Heinsfeild et al~\cite{heinsfeld2018identification}& 65.8 & 63.8 
 \\
 SVM & 59.3 & 51.7
 \\
 Random forest & 62.6 & 58.6 \\
 \hline
\end{tabular}
\end{table}

Based on the results in Table~\ref{table:acc_otherregions}, our proposed method with and without the augmentation process performs better than existing methods. Note that the classification accuracy obtained using these parcellations are below the accuracy obtained using CC-200 atlas, which implies that the pairwise correlations among CC-200 regions contain more discriminative patterns than AAL and Dosenbakh160 atlases.

\section{Conclusion and future work}\label{sec:conclusion}
In this paper, we targeted the problem on classifying subjects with ASD disorder from healthy subjects. We used fMRI data provided by ABIDE consortium, which has been collected from different brain imaging centers. No assumption or utilization of any demographic information is considered in this study. Our approach, called \emph{ASD-DiagNet}, is based on using the most correlated and anti-correlated connections of the brain as feature vectors and using an autoencoder to extract lower dimensional patterns from them. The autoencoder and a single layer perceptron are trained in a joint approach for performing feature selection and classification. We also proposed a data augmentation method in order to increase the number of samples using the available training set. We tested this method by performing 10-fold cross-validation on the whole dataset and achieved $70.1$\% accuracy in $40$ minutes. The running time of our approach is significantly shorter than $6$ hours needed by the state of the art method while achieving higher classification accuracy. In another experiment, we evaluated our method by performing 5-fold cross-validation on each data center, separately. The average result shows significant improvement in accuracy compared to the state of the art method. In this case, data augmentation helps to improve the accuracy by around $2$\%. These results demonstrate that our approach can be used for both intra-site brain imaging data, which are usually small sets generated in research centers, and bigger multi-site datasets like ABIDE in a reasonable amount of time.

\ifCLASSOPTIONcompsoc
  \section*{Funding}
\else
  \section*{Funding}
\fi
This research was supported by National Institute of General Medical Sciences (NIGMS), NIH Award Number R15GM120820, and National Science Foundations (NSF) under Award Numbers NSF CRII CCF-1464268, NSF CRII CCF- 1855441, NSF CAREER ACI-1651724 and NSF OAC 1925960. The content is solely the responsibility of the authors and does not necessarily represent the official views of governmental agencies.

\ifCLASSOPTIONcaptionsoff
  \newpage
\fi



%
{\balance
\bibliographystyle{IEEEtran}
\bibliography{References.bib}

\begin{thebibliography}{10}
\providecommand{\url}[1]{#1}
\csname url@samestyle\endcsname
\providecommand{\newblock}{\relax}
\providecommand{\bibinfo}[2]{#2}
\providecommand{\BIBentrySTDinterwordspacing}{\spaceskip=0pt\relax}
\providecommand{\BIBentryALTinterwordstretchfactor}{4}
\providecommand{\BIBentryALTinterwordspacing}{\spaceskip=\fontdimen2\font plus
\BIBentryALTinterwordstretchfactor\fontdimen3\font minus
  \fontdimen4\font\relax}
\providecommand{\BIBforeignlanguage}[2]{{%
\expandafter\ifx\csname l@#1\endcsname\relax
\typeout{** WARNING: IEEEtran.bst: No hyphenation pattern has been}%
\typeout{** loaded for the language `#1'. Using the pattern for}%
\typeout{** the default language instead.}%
\else
\language=\csname l@#1\endcsname
\fi
#2}}
\providecommand{\BIBdecl}{\relax}
\BIBdecl

\bibitem{nickel2017early}
R.~E. Nickel and L.~Huang-Storms, ``Early identification of young children with
  autism spectrum disorder,'' \emph{The Indian Journal of Pediatrics}, vol.~84,
  no.~1, pp. 53--60, 2017.

\bibitem{national2018attention}
``Attention deficit hyperactivity disorder: diagnosis and management of {ADHD}
  in children, young people and adults.''\hskip 1em plus 0.5em minus
  0.4em\relax National Collaborating Centre for Mental Health (UK), British
  Psychological Society, 2018.

\bibitem{baio2018prevalence}
J.~Baio, L.~Wiggins, D.~L. Christensen, M.~J. Maenner, J.~Daniels, Z.~Warren,
  M.~Kurzius-Spencer, W.~Zahorodny, C.~R. Rosenberg, T.~White \emph{et~al.},
  ``Prevalence of autism spectrum disorder among children aged 8 years—autism
  and developmental disabilities monitoring network, 11 sites, united states,
  2014,'' \emph{MMWR Surveillance Summaries}, vol.~67, no.~6, p.~1, 2018.

\bibitem{hosseini2016alzheimer}
E.~Hosseini-Asl, G.~Gimel'farb, and A.~El-Baz, ``Alzheimer's disease
  diagnostics by a deeply supervised adaptable 3{D} convolutional network,''
  \emph{arXiv preprint arXiv:1607.00556}, 2016.

\bibitem{eslami2018sing}
T.~Eslami and F.~Saeed, ``Similarity based classification of {ADHD} using
  singular value decomposition,'' in \emph{Proceedings of the ACM International
  Conference on Computing Frontiers 2018}.\hskip 1em plus 0.5em minus
  0.4em\relax ACM, 2018, pp. 19--25.

\bibitem{khazaee2017classification}
A.~Khazaee, A.~Ebrahimzadeh, A.~Babajani-Feremi, A.~D.~N. Initiative
  \emph{et~al.}, ``Classification of patients with {MCI} and {AD} from healthy
  controls using directed graph measures of resting-state f{MRI},''
  \emph{Behavioural brain research}, vol. 322, pp. 339--350, 2017.

\bibitem{yang2014deep}
Z.~Yang, S.~Zhong, A.~Carass, S.~H. Ying, and J.~L. Prince, ``Deep learning for
  cerebellar ataxia classification and functional score regression,'' in
  \emph{International Workshop on Machine Learning in Medical Imaging}.\hskip
  1em plus 0.5em minus 0.4em\relax Springer, 2014, pp. 68--76.

\bibitem{peng2013extreme}
X.~Peng, P.~Lin, T.~Zhang, and J.~Wang, ``Extreme learning machine-based
  classification of adhd using brain structural mri data,'' \emph{PloS one},
  vol.~8, no.~11, p. e79476, 2013.

\bibitem{colby2012insights}
J.~B. Colby, J.~D. Rudie, J.~A. Brown, P.~K. Douglas, M.~S. Cohen, and
  Z.~Shehzad, ``Insights into multimodal imaging classification of {ADHD},''
  \emph{Frontiers in systems neuroscience}, vol.~6, p.~59, 2012.

\bibitem{deshpande2015fully}
G.~Deshpande, P.~Wang, D.~Rangaprakash, and B.~Wilamowski, ``Fully connected
  cascade artificial neural network architecture for attention deficit
  hyperactivity disorder classification from functional magnetic resonance
  imaging data,'' \emph{IEEE transactions on cybernetics}, vol.~45, no.~12, pp.
  2668--2679, 2015.

\bibitem{heinsfeld2018identification}
A.~S. Heinsfeld, A.~R. Franco, R.~C. Craddock, A.~Buchweitz, and F.~Meneguzzi,
  ``Identification of autism spectrum disorder using deep learning and the
  abide dataset,'' \emph{NeuroImage: Clinical}, vol.~17, pp. 16--23, 2018.

\bibitem{craddock2013neuro}
C.~Craddock, Y.~Benhajali, C.~Chu, F.~Chouinard, A.~Evans, A.~Jakab, B.~S.
  Khundrakpam, J.~D. Lewis, Q.~Li, M.~Milham \emph{et~al.}, ``The neuro bureau
  preprocessing initiative: open sharing of preprocessed neuroimaging data and
  derivatives,'' \emph{Neuroinformatics}, 2013.

\bibitem{iidaka2015resting}
T.~Iidaka, ``Resting state functional magnetic resonance imaging and neural
  network classified autism and control,'' \emph{Cortex}, vol.~63, pp. 55--67,
  2015.

\bibitem{chen2016multivariate}
H.~Chen, X.~Duan, F.~Liu, F.~Lu, X.~Ma, Y.~Zhang, L.~Q. Uddin, and H.~Chen,
  ``Multivariate classification of autism spectrum disorder using
  frequency-specific resting-state functional connectivity—a multi-center
  study,'' \emph{Progress in Neuro-Psychopharmacology and Biological
  Psychiatry}, vol.~64, pp. 1--9, 2016.

\bibitem{abraham2017deriving}
A.~Abraham, M.~P. Milham, A.~Di~Martino, R.~C. Craddock, D.~Samaras,
  B.~Thirion, and G.~Varoquaux, ``Deriving reproducible biomarkers from
  multi-site resting-state data: An autism-based example,'' \emph{NeuroImage},
  vol. 147, pp. 736--745, 2017.

\bibitem{plitt2015functional}
M.~Plitt, K.~A. Barnes, and A.~Martin, ``Functional connectivity classification
  of autism identifies highly predictive brain features but falls short of
  biomarker standards,'' \emph{NeuroImage: Clinical}, vol.~7, pp. 359--366,
  2015.

\bibitem{parisot2018disease}
S.~Parisot, S.~I. Ktena, E.~Ferrante, M.~Lee, R.~Guerrero, B.~Glocker, and
  D.~Rueckert, ``Disease prediction using graph convolutional networks:
  Application to autism spectrum disorder and alzheimer’s disease,''
  \emph{Medical image analysis}, 2018.

\bibitem{sen2018general}
B.~Sen, N.~C. Borle, R.~Greiner, and M.~R. Brown, ``A general prediction model
  for the detection of {ADHD} and autism using structural and functional
  {MRI},'' \emph{PloS one}, vol.~13, no.~4, p. e0194856, 2018.

\bibitem{nielsen2013multisite}
J.~A. Nielsen, B.~A. Zielinski, P.~T. Fletcher, A.~L. Alexander, N.~Lange,
  E.~D. Bigler, J.~E. Lainhart, and J.~S. Anderson, ``Multisite functional
  connectivity {MRI} classification of autism: Abide results,'' \emph{Frontiers
  in human neuroscience}, vol.~7, p. 599, 2013.

\bibitem{subbaraju2017identifying}
V.~Subbaraju, M.~B. Suresh, S.~Sundaram, and S.~Narasimhan, ``Identifying
  differences in brain activities and an accurate detection of autism spectrum
  disorder using resting state functional-magnetic resonance imaging: A spatial
  filtering approach,'' \emph{Medical image analysis}, vol.~35, pp. 375--389,
  2017.

\bibitem{fredo2018diagnostic}
A.~J. Fredo, A.~Jahedi, M.~Reiter, and R.-A. M{\"u}ller, ``Diagnostic
  classification of autism using resting-state f{MRI} data and conditional
  random forest,'' \emph{Age (years)}, vol.~12, no. 2.76, pp. 6--41, 2018.

\bibitem{bi2018classification}
X.-a. Bi, Y.~Wang, Q.~Shu, Q.~Sun, and Q.~Xu, ``Classification of autism
  spectrum disorder using random support vector machine cluster,''
  \emph{Frontiers in genetics}, vol.~9, p.~18, 2018.

\bibitem{guo2017diagnosing}
X.~Guo, K.~C. Dominick, A.~A. Minai, H.~Li, C.~A. Erickson, and L.~J. Lu,
  ``Diagnosing autism spectrum disorder from brain resting-state functional
  connectivity patterns using a deep neural network with a novel feature
  selection method,'' \emph{Frontiers in neuroscience}, vol.~11, p. 460, 2017.

\bibitem{bi2018diagnosis}
X.-a. Bi, Y.~Liu, Q.~Jiang, Q.~Shu, Q.~Sun, and J.~Dai, ``The diagnosis of
  autism spectrum disorder based on the random neural network cluster,''
  \emph{Frontiers in human neuroscience}, vol.~12, p. 257, 2018.

\bibitem{brown2018connectome}
C.~J. Brown, J.~Kawahara, and G.~Hamarneh, ``Connectome priors in deep neural
  networks to predict autism,'' in \emph{Biomedical Imaging (ISBI 2018), 2018
  IEEE 15th International Symposium on}.\hskip 1em plus 0.5em minus 0.4em\relax
  IEEE, 2018, pp. 110--113.

\bibitem{dvornek2017identifying}
N.~C. Dvornek, P.~Ventola, K.~A. Pelphrey, and J.~S. Duncan, ``Identifying
  autism from resting-state f{MRI} using long short-term memory networks,'' in
  \emph{International Workshop on Machine Learning in Medical Imaging}.\hskip
  1em plus 0.5em minus 0.4em\relax Springer, 2017, pp. 362--370.

\bibitem{li2018novel}
H.~Li, N.~A. Parikh, and L.~He, ``A novel transfer learning approach to enhance
  deep neural network classification of brain functional connectomes,''
  \emph{Frontiers in neuroscience}, vol.~12, p. 491, 2018.

\bibitem{khosla20183d}
M.~Khosla, K.~Jamison, A.~Kuceyeski, and M.~Sabuncu, ``3{D} convolutional
  neural networks for classification of functional connectomes,'' \emph{arXiv
  preprint arXiv:1806.04209}, 2018.

\bibitem{lindquist2008statistical}
M.~A. Lindquist \emph{et~al.}, ``The statistical analysis of f{MRI} data,''
  \emph{Statistical science}, vol.~23, no.~4, pp. 439--464, 2008.

\bibitem{eslami2018fast}
T.~Eslami and F.~Saeed, ``{Fast-GPU-PCC}: A {GPU}-based technique to compute
  pairwise pearson’s correlation coefficients for time series data—f{MRI}
  study,'' \emph{High-throughput}, vol.~7, no.~2, p.~11, 2018.

\bibitem{craddock2012whole}
R.~C. Craddock, G.~A. James, P.~E. Holtzheimer~III, X.~P. Hu, and H.~S.
  Mayberg, ``A whole brain f{MRI} atlas generated via spatially constrained
  spectral clustering,'' \emph{Human brain mapping}, vol.~33, no.~8, pp.
  1914--1928, 2012.

\bibitem{liang2012effects}
X.~Liang, J.~Wang, C.~Yan, N.~Shu, K.~Xu, G.~Gong, and Y.~He, ``Effects of
  different correlation metrics and preprocessing factors on small-world brain
  functional networks: a resting-state functional {MRI} study,'' \emph{PloS
  one}, vol.~7, no.~3, p. e32766, 2012.

\bibitem{zhang2017hybrid}
Y.~Zhang, H.~Zhang, X.~Chen, S.-W. Lee, and D.~Shen, ``Hybrid high-order
  functional connectivity networks using resting-state functional {MRI} for
  mild cognitive impairment diagnosis,'' \emph{Scientific reports}, vol.~7,
  no.~1, p. 6530, 2017.

\bibitem{baggio2014functional}
H.-C. Baggio, R.~Sala-Llonch, B.~Segura, M.-J. Marti, F.~Valldeoriola,
  Y.~Compta, E.~Tolosa, and C.~Junqu{\'e}, ``Functional brain networks and
  cognitive deficits in parkinson's disease,'' \emph{Human brain mapping},
  vol.~35, no.~9, pp. 4620--4634, 2014.

\bibitem{raschka2017python}
S.~Raschka and V.~Mirjalili, \emph{Python machine learning}.\hskip 1em plus
  0.5em minus 0.4em\relax Packt Publishing Ltd, 2017.

\bibitem{wong2016understanding}
S.~C. Wong, A.~Gatt, V.~Stamatescu, and M.~D. McDonnell, ``Understanding data
  augmentation for classification: when to warp?'' \emph{arXiv preprint
  arXiv:1609.08764}, 2016.

\bibitem{perez2017effectiveness}
L.~Perez and J.~Wang, ``The effectiveness of data augmentation in image
  classification using deep learning,'' \emph{arXiv preprint arXiv:1712.04621},
  2017.

\bibitem{eitel2015multimodal}
A.~Eitel, J.~T. Springenberg, L.~Spinello, M.~Riedmiller, and W.~Burgard,
  ``Multimodal deep learning for robust {RGB-D} object recognition,'' in
  \emph{Intelligent Robots and Systems (IROS), 2015 IEEE/RSJ International
  Conference on}.\hskip 1em plus 0.5em minus 0.4em\relax IEEE, 2015, pp.
  681--687.

\bibitem{karpathy2014large}
A.~Karpathy, G.~Toderici, S.~Shetty, T.~Leung, R.~Sukthankar, and L.~Fei-Fei,
  ``Large-scale video classification with convolutional neural networks,'' in
  \emph{Proceedings of the IEEE conference on Computer Vision and Pattern
  Recognition}, 2014, pp. 1725--1732.

\bibitem{xu2016improved}
Y.~Xu, R.~Jia, L.~Mou, G.~Li, Y.~Chen, Y.~Lu, and Z.~Jin, ``Improved relation
  classification by deep recurrent neural networks with data augmentation,''
  \emph{arXiv preprint arXiv:1601.03651}, 2016.

\bibitem{chawla2002smote}
N.~V. Chawla, K.~W. Bowyer, L.~O. Hall, and W.~P. Kegelmeyer, ``{SMOTE}:
  synthetic minority over-sampling technique,'' \emph{Journal of artificial
  intelligence research}, vol.~16, pp. 321--357, 2002.

\bibitem{yang2004pca}
K.~Yang and C.~Shahabi, ``A {PCA}-based similarity measure for multivariate
  time series,'' in \emph{Proceedings of the 2nd ACM international workshop on
  Multimedia databases}.\hskip 1em plus 0.5em minus 0.4em\relax ACM, 2004, pp.
  65--74.

\end{thebibliography}


\begin{thebibliography}{1}

\bibitem{IEEEhowto:kopka}
H.~Kopka and P.~W. Daly, \emph{A Guide to \LaTeX}, 3rd~ed.\hskip 1em plus
  0.5em minus 0.4em\relax Harlow, England: Addison-Wesley, 1999.
\end{thebibliography}
}

%

\end{document}